\documentclass[10pt]{article} 
\usepackage[preprint]{tmlr}

\usepackage{amsmath,amsfonts,bm}

\def\eqref#1{equation~\ref{#1}}

\def\1{\bm{1}}

\DeclareMathAlphabet{\mathsfit}{\encodingdefault}{\sfdefault}{m}{sl}
\SetMathAlphabet{\mathsfit}{bold}{\encodingdefault}{\sfdefault}{bx}{n}

\usepackage{hyperref}
\usepackage{url}
\usepackage{caption}
\usepackage{subcaption}
\usepackage{graphicx}
\usepackage{multirow}
\usepackage{pifont}
\usepackage{wrapfig}

\newcommand{\yes}{\ding{52}}
\newcommand{\model}{$\texttt{Pivot}$}

\title{Leveraging Procedural Knowledge and Task Hierarchies for \\ Efficient Instructional Video Pre-training}

\author{\name Karan Samel \email ksamel@gatech.edu \\
      \addr Georgia Insitute of Technology
      \AND
      \name Nitish Sontakke \email nitishsontakke@gatech.edu \\
      \addr Georgia Insitute of Technology
      \AND
      \name Irfan Essa \email irfan@gatech.edu\\
      \addr Georgia Insitute of Technology \\
      Google DeepMind}

\def\month{MM}  
\def\year{YYYY} 
\def\openreview{\url{https://openreview.net/forum?id=XXXX}} 

\begin{document}

\maketitle

\begin{abstract}
Instructional videos provide a convenient modality to learn new tasks (ex. cooking a recipe, or assembling furniture). A viewer will want to find a corresponding video that reflects both the overall task they are interested in as well as contains the relevant steps they need to carry out the task. To perform this, an instructional video model should be capable of inferring both the tasks and the steps that occur in an input video. Doing this efficiently and in a generalizable fashion is key when compute or relevant video topics used to train this model are limited. To address these requirements we explicitly mine task hierarchies and the procedural steps associated with instructional videos. We use this prior knowledge to pre-train our model, \model, for step and task prediction. During pre-training, we also provide video augmentation and early stopping strategies to optimally identify which model to use for downstream tasks. We test this pre-trained model on task recognition, step recognition, and step prediction tasks on two downstream datasets. When pre-training data and compute are limited, we outperform previous baselines along these tasks. Therefore, leveraging prior task and step structures enables efficient training of \model{} for instructional video recommendation. 
\end{abstract}
\section{Introduction}

\begin{wrapfigure}{r}{0.5\textwidth}
  \begin{center}
\includegraphics[width=0.5\textwidth]{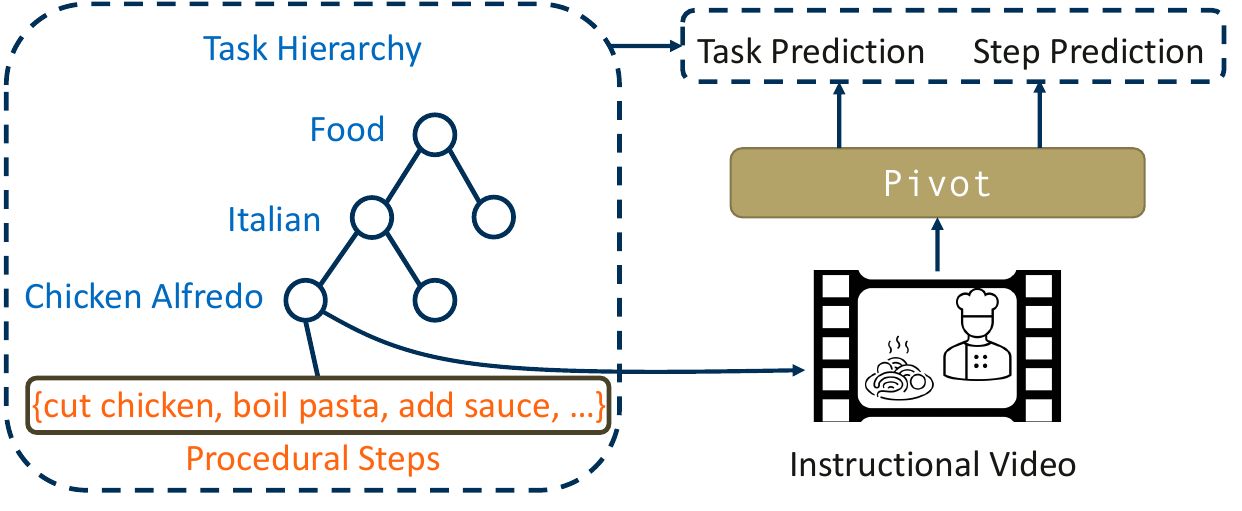}
\end{center}
\caption{We leverage both task hierarchical data as well as procedural step information to pre-train our instructional video model \model.}
\label{fig:overview}
\end{wrapfigure}

Instructional videos are a quick and convenient way to visually understand how to carry out tasks such as cooking a meal, filling out an application, or assembling furniture. A viewer then would want to find the most relevant video recommendations containing a certain cooking ingredient or a furniture fixture that they are interested in. However, the videos may not directly provide explicit annotations for these different steps. Therefore it is important that these instructional videos can be categorized within different tasks, and the distinct \emph{steps} occurring within the videos can be predicted explicitly.

Even when distinct steps within individual videos are predicted, the video recommendations must be contextualized to the user's preferences. If a user is looking up recipes with chicken, videos related to Chicken Cesar Salad or Chicken Alfredo are equally relevant. If we also know the user was looking at Italian recipes, the understanding that Chicken Alfredo pertains to Italian cuisine would be presented. Similarly, if the user was looking at low-carbohydrate meals, then the additional categorizing salads and soups videos would lead to the recommendation of the Chicken Cesar Salad. In addition to individual steps within the videos, we benefit from models that can contextualize the steps within their video-level \emph{tasks}.


Typically video representation models learn task-level representations by \emph{implicitly} learning the relationships between different steps in different tasks. We \emph{explicitly} learn these step and task relationships by mining how-to procedural steps as well as video hierarchy data, illustrated in Figure \ref{fig:overview}. For example, a video discussing Chicken Alfredo would be under the categories: Food and then Italian. A model knowing which tasks and similar steps belong to food prep and Italian cooking leverages both global task information as well as local step information. 

This leads to more efficient video pre-training for instructional fine-tuning tasks, due to this joint contextualization of video and clip-level features on prior hierarchical knowledge and procedural steps respectively. If the output video embedding aligns with other videos of Italian cooking, then it provides a strong prior on the procedural steps expected in the video (ex. cut a chicken breast, boil pasta, add Alfredo sauce, etc.) for downstream step-wise tasks related to Italian cooking. Similarly, such clips observed in a video would provide a strong prior for downstream video-level tasks for classification and clustering (ex. find similar recipes). 

By explicitly leveraging these prior procedural steps and hierarchy structure, less pre-training data is required to train the video model from scratch, beneficial in low-resource settings where time, domain-specific data, or compute is limited. We incorporate these findings with a \underline{\textbf{P}}rocedural-\underline{\textbf{H}}ierarchical \underline{\textbf{I}}ntegrated \underline{\textbf{Vi}}deo \underline{\textbf{T}}ransformer (PHIViT or \model) model, with the following contributions:
\begin{itemize}
    \item Architecture: We infer the video's task hierarchy path in addition to predicting individual steps to perform joint video-clip pre-training. 
    \item Training: The model uses different video augmentation procedures during pre-training to identify salient steps and optimal early stopping strategies. 
    \item Evaluation: The model is evaluated on video, clip, and forecasting tasks on two downstream datasets. We improve performance over previous methods and show which \model{} pre-training settings work best.
\end{itemize}
\section{Related Works}
Pre-training a video model for clip and video-level representation learning relies on a large amount of unstructured video data. These datasets, such as HowTo100M \citep{miech19howto100m} contain automatic speech recognition captions over the entire continuous video, thus they do not contain explicit steps occurring within the videos. If steps occurring with videos are discretized, it provides a method of learning a better representation of instructional videos. 

\subsection{Mining Procedural Knowledge}
Existing works aim to obtain these discrete steps directly from videos. Earlier works aim to first identify the relevant scenes of the videos by clustering video embeddings to identify the salient steps across videos to use \citep{shah2023steps}. \cite{rohrbach2022tl} extracts salient steps by finding repeat steps occurring across multiple videos and are demonstrated verbally. Similarly \cite{wang2023self} performs this using an inverse optimal transport problem between text and visual semantic embeddings. Given key steps in the video, a hypergraph can be created that links the common steps between the videos \citep{bansal2024united}. This provides a video-oriented approach to mining step-by-step procedures. Paths, consisting of chains of steps, along these graphs are used to predict the step-by-step actions required. Here smaller thus more common sub-paths are a more reliable indicator of subsequent steps than any path between two steps \citep{li2023skip}. This is further developed by \cite{zou2024weakly} through adding ordering constraints of steps. To generalize the variance of clips representing the same discrete steps, \cite{yang2023implicit} learn a model with affordance knowledge to identify equivalent entities and behaviors from HowTo100M. 

Beyond the video data itself, external sources contain step-by-step procedures for a variety of instructional tasks, such as wikiHow \citep{koupaee2018wikihow}. \cite{zhou2023non} learns how to convert a linear chain of steps to create graphs, containing steps that are interchangeable or optional. Works leveraging distant supervision (LwDS) in \cite{lwds} and in \cite{videotf} Video Taskformer (VideoTF) align the wikiHow step captions to the clip speech transcripts to provide step pseudo-labels per video clip. \cite{paprika} leverages these pseudo-labels to create a Procedural Knowledge Graph (PKG) of step labels across aligned video clips, therefore leveraging both external knowledge as well as video data to create the step-by-step instructions for HowTo100M video tasks. These discrete structures can then be used to pre-train instructional video models. 

\subsection{Procedural Pre-training}
The contrastive pre-trained model (MIL-NCE) \cite{miech19howto100m} is used to align step description embeddings to the video captions embeddings for clip-level tasks, such as step prediction. \cite{paprika} uses the PKG to learn a clip lightweight procedural-aware model (Paprika) tuned on multiple video-level and clip-level pseudo-labels. This model is then used to efficiently contextualize clip embedding inputs into a shallow Transformer \citep{vaswani2017attention} for downstream tasks. 

Instead of only clip-level inputs, the entire video context consisting of multiple clips can be used for pre-training. LwDS pre-trains a video-level Transformer to infer the distribution of step pseudo-labels for each video clip input. VideoTF builds on top of this by testing a masking objective to recover the predicted steps for random clips. These models learn video-level information implicitly through clip-step prediction objectives. With \model, we jointly learn procedural steps in addition to video-level representations explicitly within a Transformer.
\section{Method}

\begin{figure}[t]
\begin{center}
\includegraphics[width=\textwidth]{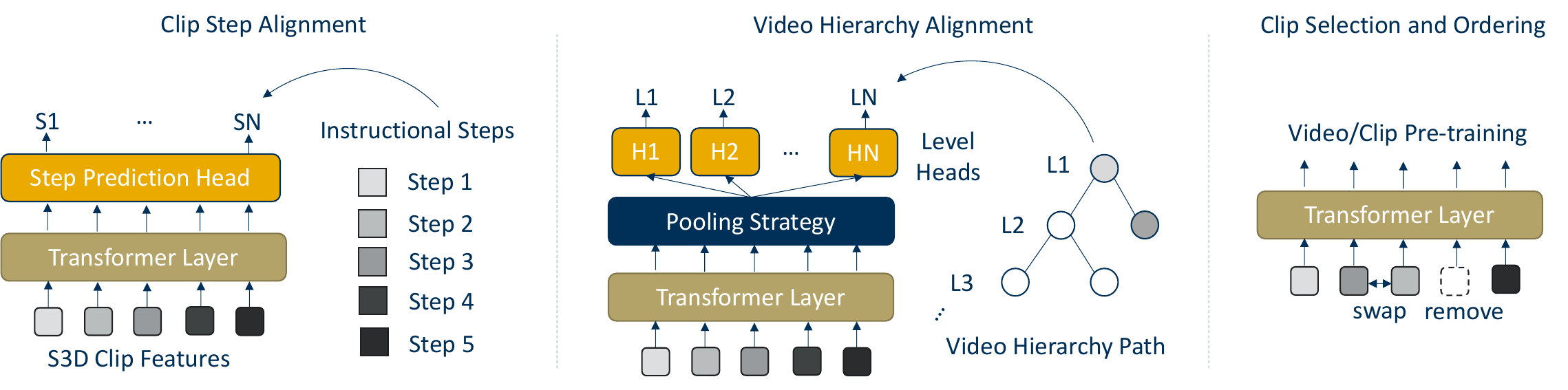}
\end{center}
\caption{Our model, \model{}, pre-trains on instructional videos to predict: a) which procedural steps belong to each video clip (left), b) where in the video hierarchy the current video belongs to (center), and c) video clip augmentation and training procedures to learn the joint clip-video representations most effectively (right).}
\end{figure}

The goal of our approach is to use both prior procedural step-level information as well as task-level hierarchy metadata to efficiently pre-train video representations for downstream instructional tasks. 
This involves inferring the clip-level step labels given a set of prior procedural steps as done in prior work (\S \ref{methods:clip_kg}). Our novel contributions involve video-level hierarchical task prediction (\S \ref{methods:video_hier}), video clip ordering and selection (\S \ref{methods:clip_sel}), and analytical early stopping strategies (\S \ref{methods:early_stop}) to efficiently pre-train a video representation model. These contributions lead to improved downstream performance in task recognition, step recognition, as well as step forecasting objectives, which are important in recommending the correct instructional videos to users.

\subsection{Clip to Procedural Steps Alignment}
\label{methods:clip_kg}

The first pre-training task involves aligning the clip-level representations with steps associated with that task occurring in the instructional video. These associated steps can be derived from the videos themselves to identify a common subset of narrated steps (ex. cut the chicken, boil pasta) along the same video task (ex. making Chicken Alfredo). Similarly, the steps required for that task can be found on a how-to tutorial website, such as an online recipe or wikiHow. For each of these instructional tasks $T \in \mathcal{T}$, we leverage procedural knowledge by using the task steps $s \in S$ provided in a sequential order $T_i = \{s_1, s_2, \dots, s_k\}$. Here each $s$ is a text description of that step. The steps may occur in more than one task $T$ and each task may have a different number of steps.

These steps then have to be aligned with clips $x$ in a video $X$, where segments for pre-training are not previously annotated. In this case, each video is segmented into 9.6 second clips. Each clip is further made out of 3.2 second segments, where 32 frames are fed into the HowTo100M pre-trained MIL-NCE video branch \citep{miech19howto100m} at 10 fps to produce each segment. Then 3 segments are mean pooled to generate each 9.6 second clip representation $X_i = \{x_1, x_2, \dots, x_n\}$, as done in \cite{paprika}. 

For the alignment, we select the top-scoring pair alignment between the clip and any steps $S$. The embeddings are obtained using MPNet \citep{mpnet}, where cosine similarity (sim($\cdot, \cdot$)) is calculated between each corresponding ASR caption $C_i = \{c_1, c_2, \dots, c_n\}$ of the 9.6 second clips and each task step in $S$. These top scoring pairs constitute pseudo-labels used per clip during pre-training $y_i = \text{top-}k\ \text{sim}(\text{MPNet}(c_i), \text{MPNet}(s))\ \forall s \in S$ corresponding to each clip $x_i \in X$. Therefore we have $y_i \in Y$ as the complete set of labels for all clips in the video, where $|Y| = |X|$. 

To train \model{} the base architecture is a standard single  Transformer encoder layer (TF-Enc). The input to the model is the entire sequence of mean pooled embeddings $X$. Then each positional output $\model(X)_i$ corresponding to clip $x_i$ is fed into a task head, which consists of a two layer MLP with a ReLU activation function in between. The task head predicts that clip's distribution over all task steps $S$ to compute a binary cross entropy (BCE) loss $\mathcal{L}_{step} = \sum_{X}\sum_{y_i \in Y}\text{BCE}(\text{MLP}(\text{TF-Enc}(X)_i), y_i)$. Recall that each $y_i$ may contain multiple step pseudo-labels, depending on our $\text{top-}k$ setting.

\subsection{Video to Hierarchy Alignment}
\label{methods:video_hier}

For the video-level objective, we predict the video's categorization \emph{path} within a larger hierarchy. These paths are categorized by users or are automatically tagged on platforms such as YouTube when a video is uploaded. We take the hierarchy paths and use our model to predict each node within the hierarchy path for that video (ex. Food $\rightarrow$ Italian $\rightarrow$ Chicken Alfredo). The model that can predict these coarse-to-fine video topics guides the step-level predictions required, and vice versa. 

We define this hierarchy as $H$ and containing nodes $n \in H$. These hierarchy nodes $n$ are associated with a single parent node and children node(s). At the first level, the hierarchy contains root node(s), which have no parents. Each level may also contain leaf nodes that have parents, but no subsequent children. Each video $X$ is already associated with a corresponding hierarchy leaf node $n_l$ at level $l \in [1, L]$ in the hierarchy, where $L$ is the max depth of the hierarchy. Therefore, each video $X$ has a corresponding path in the hierarchy $P=\{n_1, n_2, \hdots, n_l\}$. This starts with the highest level root node $n_1$ and goes down to the leaf node $n_l$ associated with the video. 


Predicting the path $P$ is done by first pooling the position-wise clip outputs $\text{TF-Enc}(X)_i$ to leverage the contextualized clip-level representations to generate a video-level embedding $v$. We test two pooling mechanisms. The first is average pooling the positional outputs $v = \frac{1}{|X|} \sum_i \text{TF-Enc}(X)_i$. The second is to stack another Transformer layer on top of the first one and use the first position classification output embedding $v = \text{TF-Enc}(\text{TF-Enc}(X))_1$. Using the single video-level embedding produced from either method, we use MLP task heads $H_i$ for each level of our hierarchy to predict the correct node at each level of the path using a cross-entropy loss (CE) $\mathcal{L}_{path} =\sum_v \sum_{n_i \in P} \text{CE}(H_i(v), n_i)$.

\subsection{Clip Selection and Ordering}
\label{methods:clip_sel}
During step prediction, \cite{lwds} and \cite{videotf} reported better results when predicting over all the steps $S$ across all tasks. However, we study clip selection in the context of our video-level alignment as well. In this context, we want to ensure that the clips associated with each hierarchy path are semantically meaningful to infer the video's topic $T$. To do so we remove and reorder the clips from the original sequence of clips by leveraging the procedural steps metadata.

\subsubsection{Clip Thresholding and Filtering}

To filter down which clips $x_i \in X$ are used, we first threshold any associated video captions $c_i \in C$ and steps $s \in S$ with a dot product above 1.0. This threshold works well in practice when using different magnitude embeddings from the captions and steps using MPNet. The new subset of clips can be defined as $X' = \{x_i \in X \mid \text{MPNet}(c_i) \cdot \text{MPNet}(s) > 1.0\}$. This improves the precision of the clips used and avoids any filler content unrelated to the video topic. 

The second filtering method only keeps clips that have step pseudo-labels belonging to the wikiHow steps of the video's associated topic $T$. To do this we find the closest topic that belongs to the video's node category. This is done as the hierarchical node names $n_l$ (Chicken Alfredo) and the external how-to task names $T$ (Grandma's Favorite Italian Alfredo) may not align. Concretely the topic is determined through
$\text{arg max}_{T \in \mathcal{T}}\ \text{sim}(\text{MPNet}(n_l), \text{MPNet}(T))$.
Then we take the complete set of pseudo-labels $Y$ containing task steps computed for the entire video $X$, as well as the steps associated with the topic $T$. From these pseudo-labels, we only keep the clips whose pseudo-label steps occur in the procedural steps of that topic $X' = \{x_i \in X \mid y_i \in T\}$. While the first level of filtering considers similar steps across all possible tasks, this further refines the selection of relevant clips for that topic. 

\subsubsection{Clip Reordering}

Instructional videos often contain sequences with long temporal gaps between a relevant sequence of clips. For example, when prepping Chicken Alfredo, one video may present cleaning and cutting the chicken at the start, while in another video this occurs right before seasoning and cooking. To efficiently learn the visual alignment between task steps and video clips, we reorder the clips chronologically using the pseudo-labels $Y$ based on the order present in the task steps $T$. We augment the reordering by randomly swapping neighboring clips with a probability of 0.15.  

The reordered clips often don't contain all the steps present in wikiHow and usually contain repeat clips of a step that occurs over a long duration. In cases with repeat steps, we experiment with randomly selecting a single step, such that the video contains unique steps.

Note that this selection and reordering augmentations are performed on the sequence of MIL-NCE input features $X$ before passing them into \model{} for pre-training. 

\subsection{Pre-training and Early Stopping}

\model{} is trained jointly with the clip-level pseudo-labels and the video-level path labels $\mathcal{L}_{joint} = \mathcal{L}_{step} + \mathcal{L}_{path}$, where the clips are filtered and reordered based on our selection strategy. The goal is to train generalized embeddings for auxiliary downstream tasks. These tasks may not share the same pre-training attributes (tasks, steps, and hierarchy) used during the pre-training procedure. In this case, we want to ensure that our model learns a generalized representation that does not overfit to our pre-training attributes.

Traditionally, we could keep track of the loss and stop the training once the improvement stops after a certain number of epochs. However, at this point, the model would have already overfit to our pre-training attributes. Instead, we observe if the model is capturing the general structure of our pre-training attributes. This is done by monitoring the model's accuracy on the clip-to-step prediction task, as these steps are more likely to overlap with downstream tasks than broader video hierarchy nodes in $H$ or how-to topics $\mathcal{T}$. 
However, we do not want to overfit on the step predictions when the downstream steps are different. Therefore we identify an inflection point of this step accuracy, up to which point the most generalizable features are learned. 

To determine this inflection point, the accuracy metric $M = \{m_1, m_2, \dots, m_{|M|}\}$ of step prediction per epoch number is recorded. From these discrete points, we first fit a polynomial $p(e) = a_0 + a_1 e + \dots + a_n e^n$ to estimate the functional form of the accuracy metric given the epoch by minimizing the least squares error: $\text{arg min}_p \sum_{e=1}^{|M|} |p(e) - m_e|^2$. In practice, we set our $n=10$. Given this function, we compute the first derivative and find the epoch that leads to the local maximum $p$: $e^* = \text{max}_{e \in [1, K]}\ p'(e)$. The model from this epoch $e^*$ is used for downstream fine-tuning. 


\label{methods:early_stop}
\section{Experimental Setup}

\subsection{Pre-training Setup}
We pre-train our model on the HowTo100M dataset, where we use a subset of the data used in \cite{bertasius2021space}. Out of those videos, 30k videos were still available to download, which we use in all of our baselines, experiments, and ablations. 

To obtain the tasks $\mathcal{T}$ and instructional steps $S$, we use the wikiHow corpus \citep{koupaee2018wikihow}. We select the articles following \cite{paprika}, where each article contains a task name and contains high level step descriptions. 

For video hierarchy matching, we extract the hierarchy paths from the HowTo100M dataset \citep{miech19howto100m}, where each video is classified within a hierarchy $L = 3$ levels deep. This consists of the root nodes, child nodes which are subcategories of the root node, and task (leaf) nodes $n_l$ which are subcategories of the child nodes. The root, child, and task nodes consisted of 17, 105, and 1059 nodes respectively. These are also the corresponding output prediction sizes for our MLP classifier heads $H_1, H_2$, and $H_3$ when training our $\mathcal{L}_{path}$ objective.

We jointly train the video and clip level objectives given the clip selection and ordering augmentations $\mathcal{L}_{joint} = \mathcal{L}_{step} + \mathcal{L}_{path}$. We used an Adam optimizer \citep{kingma2014adam} with a learning rate of $1e^{-4}$, a decay of $1e^{-3}$, and a batch size of 256. For our proposed method, pre-training for 2000 epochs took 27 hours using four A40 GPUs and 16 CPU threads. The clip-to-step metric was tracked to determine an early stopping checkpoint. The epoch $e^*$ that maximized the first derivative function was used for downstream fine-tuning. 

\subsection{Fine-tuning Setup}

We evaluate our pre-training setup on two datasets: COIN \citep{tang2019coin} and CrossTask \citep{zhukov2019cross}. COIN contains 11.8k videos across 180 tasks arranged in a hierarchical fashion, where the hierarchy was manually curated (different from HowTo100M). Each clip within the video is annotated with a step belonging to the video task. CrossTask contains videos across 83 tasks. For evaluation purposes, we tested 18 subtasks with clip-level step annotations with 2.3k videos. 

For each dataset, we evaluate over three tasks:
\begin{itemize}
    \item \textbf{Task Recognition}: Predict the task label from the video using the pooled video-level outputs $v$.
    \item \textbf{Step Recognition}: Predict the step associated with each clip, where the position-wise clip embeddings from the base Transformer layer are fed directly into a task head for prediction.
    \item \textbf{Step Forecasting}: Predict the step for the clip, where the input clip embedding is masked. We ensure that there is always at least one prior clip in the forecasting history.
\end{itemize}   

Each dataset and task is split for training and evaluation following \cite{paprika}. Note that the task heads used in fine-tuning are newly initialized since they cover different tasks than those in pre-training. The Adam optimizer settings are kept the same as in pre-training, with a reduced batch size of 16.


\subsection{Baseline Methods}

We compare our pre-training approach against several baselines. The first baseline is observing the performance using the original HowTo100M MIL-NCE embeddings \citep{miech19howto100m} in our architecture without any pre-training. 

We also test Paprika \citep{paprika}, which leverages the relationships between steps across different tasks within its mined procedural knowledge graph as pseudo-labels. These pseudo-labels are used to define four different objectives to train a lightweight MLP Procedure-Aware Model on individual clip inputs during pre-training. This MLP model is used to contextualize the input clip embeddings before feeding them into the Transformer for the downstream tasks.

Transformer-based baselines are also trained with LwDS \citep{lwds}, which also performs clip step prediction. For the output labels, it computes the similarity between the clip and all wikiHow steps $S$. It tests two different label strategies: taking the top-3 scoring discrete labels for step classification (SC) and continuous probability labels for distribution matching (DM). For fine-tuning it uses a second layer Transformer to pool the clip representations for inference. 

Another method is VideoTF \citep{videotf}, which builds on top of LwDS by inferring \emph{masked} clip labels instead of inferring all clip labels unmasked inputs. During pre-training it uses a two layer Transformer for step prediction. During fine-tuning it further tunes the corresponding task heads for downstream evaluation. 

All baselines used the same pre-training and fine-tuning optimization settings as our method. For pre-training, each baseline was trained for 2000 epochs. We refer back to the original papers for further implementation details regarding these baselines.

\section{Results}

\subsection{Baseline Performance}
\begin{table}[t]
    \centering
    \caption{Percentage accuracy for task recognition (TR), step recognition (SR), and step forecasting (SF) within the COIN and CrossTask datasets across different pre-training methods. All methods were pre-trained on the same subset of 30k videos. Best performances are in \textbf{bold} while runner-ups are \underline{underlined}.}
    \begin{tabular}{c c c c|c c c|c c c}
        \multirow{2}{*}{Method} & Pre-Training & Task & Step & \multicolumn{3}{|c|}{COIN} & \multicolumn{3}{|c}{CrossTask} \\
        & Architecture & Loss & Loss & SF & SR & TR & SF & SR & TR \\
        \hline
        MIL-NCE & - & & & 36.87 & 38.73 & 78.30 & 58.07 & 57.70 & 89.43 \\
        Paprika & MLP & \yes & \yes & \underline{42.54} & 45.57 & 84.40 & 59.49 & 60.01 & 93.90 \\
        \hline
        LwDS (SC) & \multirow{4}{*}{Transformer} & & \yes & 37.05 & 43.42 & 81.02 & 59.49 & 59.84 & 92.07 \\
        LwDS (DM) & & & \yes & 40.60 & \underline{48.41} & 85.39 & 61.04 & 61.31 & 94.31 \\
        VideoTF (SC) & & & \yes & 37.74 & 42.05 & 84.12 & 57.95 & 57.58 & 94.11 \\
        VideoTF (DM) & & & \yes & 37.35 & 38.08 & 84.87 & 59.28 & 60.06 & \textbf{94.92} \\
        \hline
        \multirow{2}{*}{\model{} (Ours)} & \multirow{2}{*}{Transformer} & \yes & & 42.31 & 46.78 & \underline{86.70} & \textbf{62.18} & \underline{62.20} & 94.72 \\
        & & \yes & \yes & \textbf{42.68} & \textbf{49.89} & \textbf{87.42} & \underline{62.00} & \textbf{62.62} & \textbf{94.92} \\
    \end{tabular}
    \label{tab:baselines}
\end{table}
\begin{table}[t]
    \centering
    \caption{The model performances across different pre-training dataset sizes are compared.}
    \begin{tabular}{c c|c c c|c c}
        \multirow{2}{*}{Method} & Pre-Training & \multicolumn{3}{|c|}{COIN} & \multicolumn{2}{|c}{CrossTask} \\
        & Videos & SF & SR & TR & SF & SR \\
        \hline
        \model{} & 30k & \underline{42.68} & 49.89 & 87.42 & \underline{62.00} &\underline{62.62} \\
        Paprika & 85k & 42.65 & 50.48 & 85.31 & 61.42 & 62.38 \\
        \hline
        Paprika & \multirow{3}{*}{1.2M} & \textbf{43.22} & \underline{50.99} & 85.84 & \textbf{62.63} & \textbf{63.53} \\
        LwDS (DM) & & 39.4 & 54.1 & \underline{90.0} & - & - \\
        VideoTF (DM) & & 42.4 & \textbf{54.8} & \textbf{91.0} & - & - \\
    \end{tabular}
    \label{tab:data_efficiency}
\end{table}
We compare our method against previous works in Table \ref{tab:baselines}. Without any task pre-training, MIL-NCE provides us with a baseline performance for both COIN and CrossTask. Paprika substantially improves upon this basis using only a shallow pre-trained embedding model. This further demonstrates that leveraging prior knowledge within the PKG and well defined objectives leads to efficient pre-training. With Transformer-based pre-trained models, LwDS improves upon Paprika using the distribution matching strategy over all step labels. In comparison, VideoTF had lower performance due to its masking learning strategy over our smaller basis of 30k videos. 

We test our framework initially with just the video hierarchy path alignment objective $\mathcal{L}_{path}$ using the Transformer pooling layer to compute the video level embedding $v$. In this setting, we see improvements over the baseline results, even when we do not leverage the clip-level labels. This indicates that the right model architecture and video-level hierarchical data provides the largest improvement in downstream model performance. 

When we use our full model with clip step alignment, we improve the performance further. For the augmentations we: 1) keep clips with dot product similarities above 1.0, 2) are part of the linked task steps, and 3) are ordered by the step order. We ablate these choices in \S \ref{sec:ablations}.

To understand the data efficiency of \model{} we report the performances of the baseline models given their original pre-training dataset sizes in Table \ref{tab:data_efficiency}. 
Compared to Paprika with 85k samples, \model{} outperforms it with 30k samples. While using the full 1.2M samples, \model{} still has comparable results, with strong task recognition performance. For LwDS and VideoTF reported with 1.2M samples, improvements are seen in the step and task recognition methods when simpler step prediction objectives have access to more pre-training data. An interesting observation is that by leveraging task-level objectives, \model{} and Paprika add significant gains in forecasting tasks. This is true even when step-objective only LwDS and VideoTF methods leverage the full data.

\subsection{Method Ablations}
\label{sec:ablations}
\begin{table}[t]
    \centering
    \caption{Each video input augmentation is tested for \model{} where the relevant step-clips are selected by a threshold (Thresh), belong the video task (In Task), are in order by the how-to steps (Sort), are unique steps (Unique), and are randomly swapped with neighboring steps (Swap).}
    \begin{tabular}{c c c c c c|c c c|c c c}
        \multirow{2}{*}{Pooling} & \multirow{2}{*}{Thresh} & \multirow{2}{*}{In Task} & \multirow{2}{*}{Sort} & \multirow{2}{*}{Unique} & \multirow{2}{*}{Swap} & \multicolumn{3}{|c|}{COIN} & \multicolumn{3}{|c}{CrossTask} \\
        & & & & & & SF & SR & TR & SF & SR & TR \\
        \hline
        \multirow{3}{*}{Mean} & \yes & & & & & 37.24 & 49.05 & 87.02 & 62.28 & 62.43 & 94.11\\
        & \yes & \yes & & & & 36.68 & 49.66 & 86.9 & \underline{62.62} & 62.71 & 94.31\\
        & \yes & \yes & \yes & & & 36.89 & 49.82 & 86.94 & 62.83 & \underline{62.79} & 94.11\\
        \hline
        \multirow{5}{*}{TF-Enc} & \yes & & & & & 41.59 & \underline{49.98} & \textbf{87.53} & 62.21 & 62.68 & \underline{94.31} \\
        & \yes & \yes & & & & 41.77 & \textbf{50.25} & 86.90 & 62.49 & 62.65 & 93.5 \\
        & \yes & \yes & \yes & & & \textbf{42.68} & 49.89 & \underline{87.42} & 62.0 & 62.62 & \textbf{94.92} \\
        & \yes & \yes & \yes & \yes & & \underline{42.28} & 49.82 & 86.54 & 62.18 & 62.68 & 93.7\\
        & \yes & \yes & \yes & & \yes & 41.81 & 49.57 & 86.38 & \underline{62.62} & \textbf{63.29} & 92.89\\
        & \yes & \yes & \yes & \yes & \yes & 41.33 & 49.82 & 86.82 & \textbf{62.65} & 62.82 & 93.7\\
        \hline
    \end{tabular}
    \label{tab:ablations}
\end{table}

We ablate the design choices used for our task and clip-level objectives in Table \ref{tab:ablations}. For the task-level objective, we ablate using a mean or a Transformer encoder (TF-Enc) pooler to predict the hierarchy path steps. For clip-level designs, we first test thresholding clips whose MPNet dot product between the caption top step instruction is greater than 1.0. We then test filtering steps that only belong to that video's task. The clips can also be sorted by the order in which the steps appear in the task wikiHow. Unique clips based on the step labels can also be chosen, and finally, neighboring clips can be randomly swapped to increase the input variance order of the steps per video. 

From the task-level pooling, using a second Transformer layer to compute a contextualized video embedding improves downstream performance using the same clip-level settings. 

For clip-level performance, just thresholding the clips already provides comparable performance to other clip ablation setups. This shows that even a minimum filtering effort of the video segments used during pre-training can have a large impact on our joint video-clip training. Note that the setting without thresholding would be the same setup as the LwDS baseline. Using thresholding, task steps and sorting provides similar performance to thresholding only. However, this setup speeds up pre-training by 2-3x due to the fewer clip samples required per video input, thus larger batch sizes available (256 versus 92). We observe that adding unique steps and swapping clips leads to diminishing returns. We hypothesize that this is due to drastic reduction in the number of clips per video, where the variance in the steps and their natural order in videos is lost. 

\subsection{Early Stopping Results}

\begin{table}[t]
    \centering
    \caption{We test using our analytical early stopping epoch $e^*$ across all baseline methods. The difference values are calculated by subtracting the pre-trained model saved at the last epoch (2000) from the optimal stopping epoch (accuracy epoch $e^*$ - accuracy epoch 2000).}
    \scriptsize
    \begin{tabular}{c c|c c c|c c c}
        \multirow{2}{*}{Method} & Stop & \multicolumn{3}{|c|}{COIN} & \multicolumn{3}{|c}{CrossTask} \\
        & Epoch & SF & SR & TR & SF & SR & TR \\
        \hline
        Paprika & 400 & 41.61 (-0.93) & 42.82 (-2.75) & 82.53 (-1.87) & 56.46 (-3.03) & 55.89 (-4.12) & 93.7 (-0.20) \\
        LwDS (SC) & 1000 & 37.77 (+0.72) & 43.05 (-0.37) & 81.54 (+0.52) & 59.18 (-0.31) & 60.4 (+0.56) & 91.67 (-0.40) \\
        LwDS (DM) & 200 & \underline{39.24} (-1.36) & \underline{44.39} (-4.02) & \underline{86.98} (+1.59) & \textbf{62.12} (+1.08) & \textbf{62.62} (+1.31) & 94.11 (-0.20)\\
        VideoTF (SC) & 800 & 37.75 (+0.01) & 43.46 (+1.41) & 83.8 (-0.32) & 60.11 (+2.16) & 60.81 (+3.23) & 94.11 (+0.00) \\
        VideoTF (DM) & 500 & 38.01 (+0.66) & 42.7 (+4.62) & 84.52 (-0.35) & 59.12 (-0.16) & 60.26 (+0.20) & \underline{94.72} (-0.20) \\
        \hline
        \model{} & 150 & \textbf{42.68} (+0.18) & \textbf{49.89} (+2.17) & \textbf{87.42} (+0.77) & 62.00 (-0.65) & \textbf{62.62} (+0.53) & \textbf{94.92} (+0.20)
    \end{tabular}
    \label{tab:early_stopping}
\end{table}
\begin{figure}[t]
\centering
\begin{subfigure}{.485\textwidth}
  \centering
  \includegraphics[width=\textwidth]{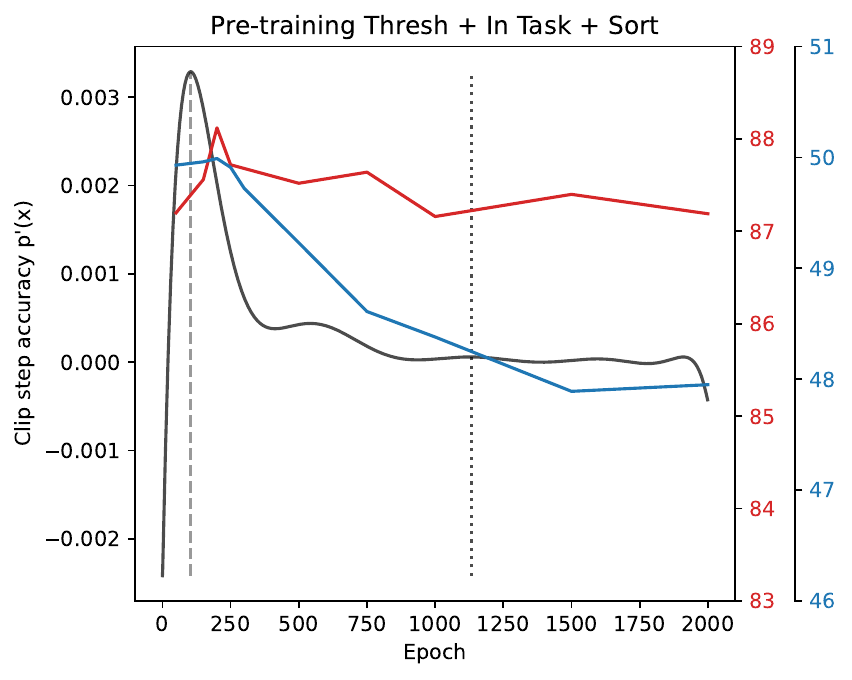}
  \label{fig:sub1}
\end{subfigure}%
\begin{subfigure}{.515\textwidth}
  \centering
  \includegraphics[width=\textwidth]{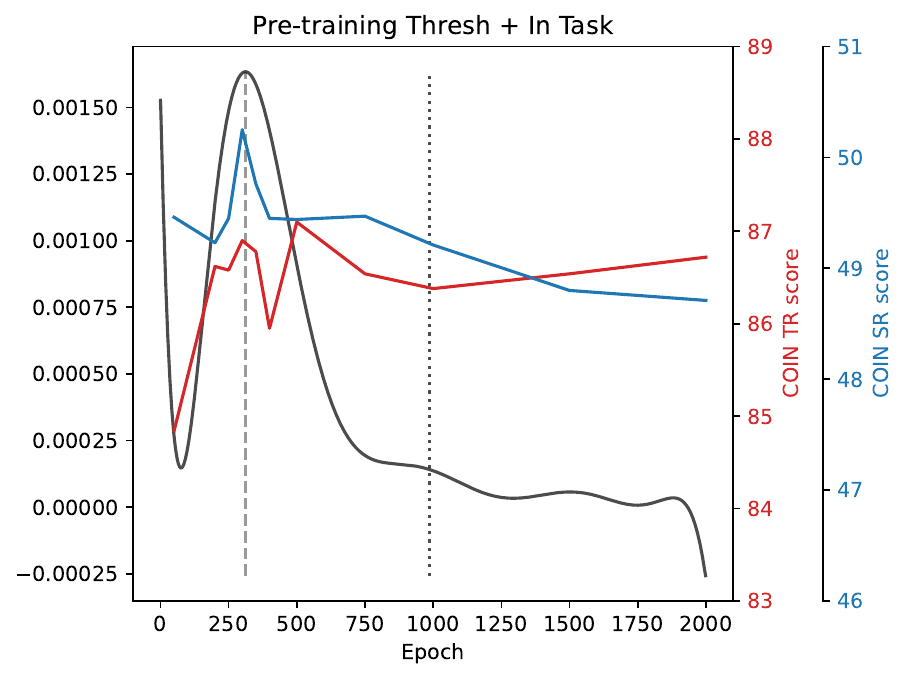}
  \label{fig:sub2}
\end{subfigure}

\bigskip
\begin{subfigure}{\textwidth}
  \centering
  \includegraphics[width=0.7\textwidth]{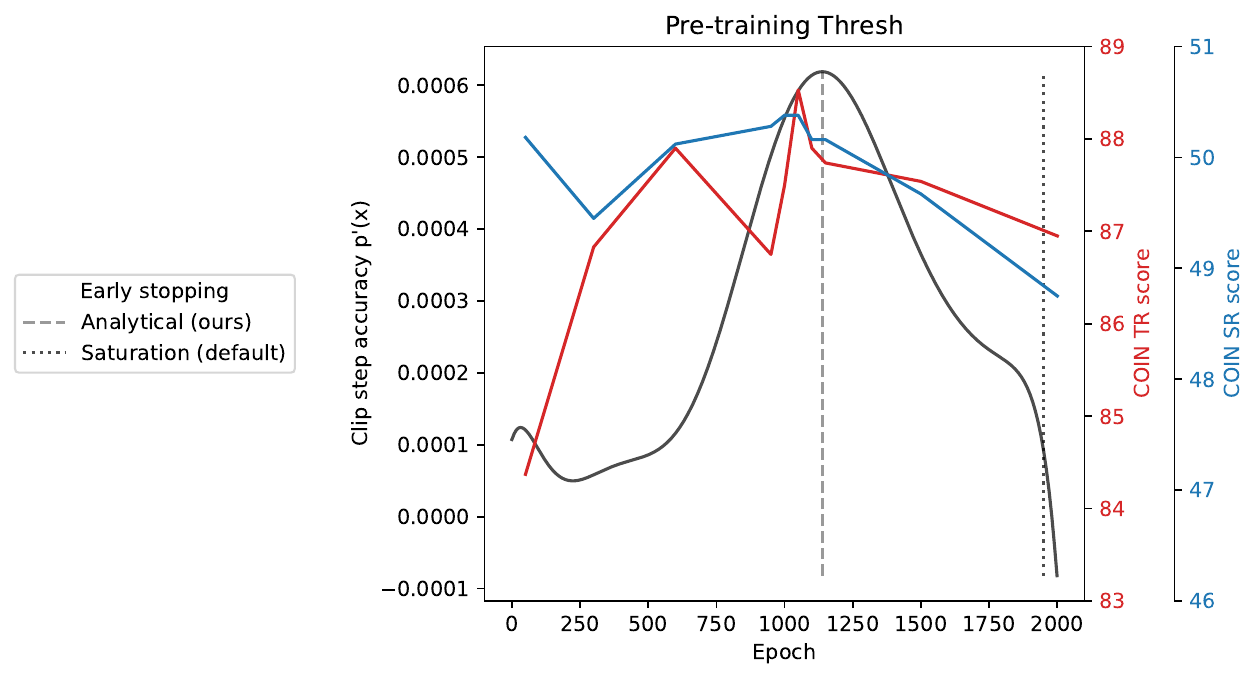}
  \label{fig:sub3}
\end{subfigure}
\caption{Pre-trained models from different epochs are tested on downstream COIN task recognition (red line) and step recognition (blue line) tasks. The derivative of the clip step accuracy $p'(x)$ is also plotted (black line), where the max value represents the analytical early stopping point (dashed line). The saturation based early stopping with no improvements over 50 epochs is also presented as a reference (dotted line). 
}
\label{fig:early_stopping}
\end{figure}

During the evaluation of our method, we took the early stopping pre-training checkpoint associated with the clip step prediction accuracy. We also test using the same early stopping approach for the baseline methods and report them in Table \ref{tab:early_stopping}. Here the average accuracy performance change from epoch 2000 to our analytical stopping checkpoint is -0.30 across all baseline methods and a change of 0.53 for our \model{} method. This indicates that the downstream performance is preserved even when running on average for 508 out of 2000 epochs. Note that during our experiments we saved the model checkpoint every 50 epochs, while the epoch metrics were kept every epoch. 


We also visualize the early stopping criterion versus downstream performance of COIN step and task recognition experiments in Figure \ref{fig:early_stopping}. The general trend across different pre-training setups is that downstream performance peaks around a certain epoch. This epoch is typically closer to our analytically computed early stopping point than a typical saturation-based stopping, which is determined by no step accuracy improvement over 50 epochs. We can observe here that in our post hoc evaluation of neighboring epochs, that optimal checkpoints may vary but remain close to the analytical epoch, under different pre-training settings. During our experimentation, we selected the saved checkpoint closest to this optimal epoch.

Here it is also shown that using our selected method of thresholding, in task, and sorting methods for clip tasks also have a faster convergence time while providing comparable performance to thresholding only augmentations. This adds to our previous motivation of using fewer clips during pre-training, to allow for \emph{faster training and faster stopping}. Removing the sorting leads to more variance for the model to learn, and doubles the epochs required. Using only thresholding requires the greatest number of epochs for convergence since each video may contain steps from different tasks and in different orders. In general, we show that it is possible to identify optimal stopping points when learning clip-level representations, thereby reducing computational costs.

\section{Conclusion}

Instructional videos provide viewers with a convenient way of learning new tasks based on their interests and the materials they have available to carry out the task. Therefore it is important to identify these video topics and the intermediate steps that pertain to them. We leverage two levels of prior knowledge through these how-to steps corresponding to videos in addition to the video's task in a larger task hierarchy structure. Leveraging these explicit knowledge structures allows our model to pre-train across different video topics more efficiently. This differs from implicitly learning task-level representations from individual clips, which requires more data, time, and compute to pre-train.

This is demonstrated by pre-training \model{} on 30k instructional videos and testing its capabilities on three downstream tasks involving task recognition, step recognition, and step forecasting across two different datasets. In these settings, \model{} outperforms previous procedural pre-training methods as it efficiently incorporates step and task-level supervision within a Transformer encoder. We also present practical pre-training augmentation strategies as well as early stopping analysis to improve the compute as well as performance efficiencies of our pre-training method. With this work, we push to further understand how to mine, and leverage structured data within models of complex modalities, such as video, in a generalizable manner.

\bibliography{main}
\bibliographystyle{tmlr}

\end{document}